\documentclass{article}

\usepackage{microtype}
\usepackage{graphicx}
\usepackage{subfigure}
\usepackage{enumitem}
\usepackage{amsfonts}
\usepackage{amsmath}
\usepackage{booktabs} 

\DeclareMathOperator{\score}{score}
\DeclareMathOperator{\nn}{NN}
\DeclareMathOperator{\margin}{margin}

\usepackage{hyperref}

\usepackage[accepted]{icml2020}

\icmltitlerunning{Contextual Lensing of Universal Sentence Representations}

\begin{document}

\twocolumn[
\icmltitle{Contextual Lensing of Universal Sentence Representations}




\begin{icmlauthorlist}
\icmlauthor{Jamie Kiros}{goo}
\end{icmlauthorlist}

\icmlaffiliation{goo}{Google Research, Toronto, Ontario, Canada}

\icmlcorrespondingauthor{Jamie Kiros}{kiros@google.com}

\icmlkeywords{Machine Learning, ICML, Natural Language Processing}

\vskip 0.3in
]



\printAffiliationsAndNotice{}  

\begin{abstract}
What makes a universal sentence encoder universal? The notion of a generic encoder of text appears to be at odds with the inherent contextualization and non-permanence of language use in a dynamic world. However, mapping sentences into generic fixed-length vectors for downstream similarity and retrieval tasks has been fruitful, particularly for multilingual applications. How do we manage this dilemma? In this work we propose Contextual Lensing, a methodology for inducing context-oriented universal sentence vectors. We break the construction of universal sentence vectors into a core, variable length, sentence matrix representation equipped with an adaptable `lens' from which fixed-length vectors can be induced as a function of the lens context. We show that it is possible to focus notions of language similarity into a small number of lens parameters given a core universal matrix representation. For example, we demonstrate the ability to encode translation similarity of sentences across several languages into a single weight matrix, even when the core encoder has not seen parallel data.
\end{abstract}

\section{Introduction}
\label{intro}

This work introduces a new framework from which to induce and use universal sentence vectors (USVs). Specifically, we focus on learning mappings of sentences into fixed-length vector representations applicable to a variety of downstream tasks focused on textual similarity, retrieval and analysis. 

Recent work has been dominated by research on large-scale self-supervised pre-training and subsequent task-specific fine-tuning \citep{dai2015semi, howard2018universal, radford2018improving, devlin2018bert}. This pretrain-finetune paradigm has led to significant improvements in a wide range of NLP tasks, primarily static classification benchmarks. While early work in USVs had evaluated their utility on such tasks, task-specific fine-tuning has largely made this application of USVs less relevant. However, there are still plenty of use cases and applications where having USVs is appealing. One of the most successful examples of USVs has been LASER \citep{artetxe2019massively}, a tool for learning language-agnostic sentence representations. LASER and related methods can be used to encode sentences across dozens of languages and has been utilized for mining parallel corpora \citep{artetxe2018margin, guo2018effective} and the creation of new parallel datasets for machine translation \citep{schwenk2019ccmatrix}. Language agnostic USVs have also been used to analyze large-scale sentential similarity graphs across languages \citep{schwenk2019analysis}. These and related tasks do not lend themselves naturally to the current pretrain-finetune paradigm on static benchmarks. The ability to flexibly encode language into vector representations often lends itself to creative and unconventional applications, as has been done in the space of word representations.

A problem arises, however, with respect to the `universal' part of universal sentence vectors. What does it mean for a sentence vector to be `universal'? Initial constructions of USVs, such as skip-thoughts \citep{kiros-nips-2015}, were motivated by the success of transfer learning in computer vision. Namely, that a single task and a high-capacity encoder could result in a generic, internal representation of images that could then be adapted to arbitrary downstream tasks. However, empirical evidence has demonstrated that this setup is substantially less effective for language than vision compared to the recent pretrain-finetune paradigm. \footnote{A trade-off here is that self-supervised learning has had a substantially higher impact in language processing than vision, with respect to supervised benchmarks.}.

This issue of universality is rooted in the inherent contextualization of language. In a dynamic, impermanent world, language as a tool to convey meaning necessarily requires context from which it can be interpreted \footnote{This work uses the term `context' very loosely to mean the influence of some outside information on meaning and similarity.}. The contexts of language use are infinite in nature, may or many not be accessible from text alone, and often have an implicit aspect to them as a function of the conversational or world scope. A USV seems to be at odds with this. If similarity is derived exclusively from the embedding of the sentence, how could this capture all the seemingly, infinite contexts in which this sentence can be interpreted? 

In this paper we propose a solution to this dilemma, called Contextual Lensing (CL). Contextual Lensing breaks up the construction of a USV into two components: a `universal' component, as a function of a large self-supervised model and a `lens' from which this universal representation is to be focused on as a function of context. Thus, a Lensed Universal Sentence Vector combines a generic variable-length \textit{matrix} representation of a sentence with a lens from which a fixed-length vector representation can be derived. As a proof of concept, we focus on arguably the simplest instantiation of this framework. That is to use a contextualized word embedding matrix from BERT \citep{devlin2018bert} or other Muppet models as the universal matrix representation, and to learn a reduction operator with `lens parameters' as a function of context to map this representation into a fixed-length vector. We consider \textit{Static Lensing}, where the reduction operator is constructed using supervised learning. This is in contrast to \textit{Dynamic Lensing}, where the lens parameters are adapted online, perhaps via FiLM \citep{perez2018film} or hyper-networks \citep{ha2016hypernetworks} in settings outside direct supervised learning, such as Meta, Few-Shot or Reinforcement Learning. We leave these considerations for future work.

\subsection{Additional Motivation}

\begin{table*}[t]
  \medskip
  \small
  \centering
  \begin{tabular}{l c c c c c}
  \toprule
  Name & Contextualized Embeddings & Encoder & Lens Data & Training Module & Dimensionality \\
  \midrule
  BERT BOW & BERT-Large & MeanPool & & & 1024 \\
  sCL(BERT-Large; NLI) & BERT-Large & Simple & NLI & Classifier & 1024 \\
  \midrule
  mBERT BOW & mBERT & MeanPool & & & 768 \\
  CL(mBERT; NLI)  & mBERT & GatedConv & NLI & Classifier & 1024 \\
  \midrule
  CL(mBERT; $N$) & mBERT & GatedConv & Translate & Ranker & 1024 \\
  sCL(mBERT; 100M) & mBERT & Simple & Translate & Ranker & 1024 \\
  CL(mBERT; 100M)$\dagger$ & mBERT & GatedConv & Translate & Ranker & 4096 \\
  sCL(mBERT; 100M)$\dagger$ & mBERT & Simple & Translate & Ranker & 4096 \\
  \bottomrule
  \end{tabular}
  
  \caption{List of contextualized USVs trained and evaluated in this work. The first group are English models. The second group are unsupervised multilingual models (no parallel data). The third group are translation models with varying numbers of parallel examples $N$=\{10k, 100k, 1M, 10M, 100M\}. Dimensionality is the final sentence vector representation size.}
  \label{tab:usrs}
\end{table*}

Before elaborating on the details of our model, we provide three additional motivations for Contextual Lensing under the instantiation of contextual embeddings + reduction:

\noindent \textbf{Ad-hoc Categories and Adaptation.} Consider an example of an ad-hoc category \citep{barsalou1983ad}. Given the following concepts: \{pets, children, parents, works of art, explosive material\}, how are they collectively related? A USV may assign a high relatedness score to children and parents but is unlikely to assign a high score to children and explosive material. Now suppose one was given the following information: 'Things to pay attention to if there is a fire'. Conditioned on this new information, the relationship between the concepts becomes apparent almost immediately.

How could a neural network model this behaviour? Notice that the underlying meanings of each concept do not change before and after the new information is introduced. Instead, only the `lens' from which we view the concepts changes. Under the lens introduced by the additional categorization, the relatedness of concepts becomes clear. Furthermore, to humans, this process happens almost instantaneously. It is unlikely that a neural network should have to fine-tune the representations of each concept as a function of the contextual category. Instead, this behaviour is better modelled as a modulation. The scores of each concept become modulated by the representation of the category. This example highlights two points: one, the existence of a universal core model that captures the underlying meaning(s) of the concepts and two, the existence of a fast, adaptive modulator that operates on the core model. This aligns with our instantiation of Contextual Lensing, albeit a simplified version.

\noindent \textbf{Self-supervision as a Foundation.} Large-scale self-supervised models provide a foundation from which additional operations can be applied. Empirical results demonstrate the effectiveness of predicting properties about the scope of the model as a means of learning useful internal representations of the world. This principle can also be applied with multiple modalities, such as the addition of images e.g. \citep{lu2019vilbert} as part of self-supervised learning with text. Models such as KERMIT \citep{chan2019kermit, chan2019mkermit} based on the Insertion Transformer framework \citep{stern2019insertion} are capable of modelling the joint distribution and all its factorizations under a single, unified self-supervised model. These have the potential to provide an excellent foundation from which USVs can be induced through Contextual Lensing.

\noindent \textbf{Context Diversity.} Language can be analyzed through seemingly infinitely many lenses. Some examples include semantic, syntactic, temporal, sociocultural and ad-hoc relatedness. Each of these can provide a new lens from which USVs can be induced. Such tools might be useful for e.g. retrieving semantically or syntactically similar sentences, inducing representations as a function of time and quickly adapting to ad-hoc categories. Furthermore, these mechanisms could potentially lead to useful tools for social scientists and other disciplines for interactive textual analysis.

\subsection{Experimental Contributions}

\begin{itemize}[leftmargin=*]
    \item We lens BERT representations to the task of Natural Language Inference (NLI) for downstream English tasks. Our results are nearly as strong as SBERT \citep{reimers2019sentence} but without any fine-tuning of BERT. Our results outperform on average all other existing USVs on 9 downstream tasks.
    \item We show it is possible to learn massively multilingual sentence vectors while encoding translation similarity into a single weight matrix.
    \item We perform an extensive comparison to LASER on 100+ languages, demonstrating improved sentence matching performance on very low resource languages.
\end{itemize}

\section{Related Work}

\noindent \textbf{Universal Sentence Vectors.} Approaches to learn USVs can be classified into unsupervised or self-supervised methods \citep{kiros-nips-2015, hill-naacl-2016, arora-iclr-2017, logeswaran-iclr-2018}, supervised \citep{conneau-emnlp-2017, reimers2019sentence}, multi-task \citep{cer-arxiv-2018, subramanian-iclr-2018} multi-lingual \citep{artetxe2019massively} and even random encoders \citep{wieting2019no}. The LASER method of \citet{artetxe2019massively} learns vectors through a sequence-to-sequence model by predicting a translated sentence from a source sentence. LASER and SBERT \citep{reimers2019sentence} are the most closely aligned with our work.

\noindent \textbf{Contextualized word embeddings.} Our method takes advantage of pre-computed contextualized word representations \citep{melamud-conll-2016, mccann-nips-2017, peters-arxiv-2018}. Initially these methods were used to augment or replace non-contextualized word embeddings. Nowadays it is more common to fine-tune the entire pipeline. Our work provides evidence that contextualized word representations from large-scale self-supervised models provide a foundation for learning sentence vectors.

\noindent \textbf{Adaptor methods.} Our proposed setup may also fall into a category of adaptor methods that aim to efficiently adapt pre-trained models to new tasks \citep{houlsby2019parameter, stickland2019bert} including few and zero-shot settings \citep{bansal2019learning, puri2019zero}. The key difference between these methods and ours is the setting. While these methods focus on adaption to new classification tasks, we focus on adapting sentence representations to various contexts and arbitrary downstream tasks.

\section{Method}

We begin by defining notation. Let $S = w^1, \ldots, w^T$ denote a length $T$ sequence according to a pre-defined subword vocabulary. Our goal is to map $S$ into a $D$-dimensional vector representation conditioned on two parameters sets: the base parameters $\Theta_B$ and the lens parameters $\Theta_L$. The base parameters are obtained from a pre-trained, self-supervised model such as BERT, while the lens parameters are learned as a function of some pre-defined context. We will describe how the context is defined and the lens parameters learned in a subsequent section. The final sentence embedding $s$ for $S$ is a composition of two functions $\Psi_B$ and $\Psi_L$ given by $s = \Psi_L(\Psi_B(S; \Theta_B); \Theta_L)$. The mapping $\Psi_B$ returns a variable-length matrix $E \in \mathbb{R}^{K \times T}$ as the universal sentence representation of $S$. The sentence encoder $\Psi_L: \mathbb{R}^{K \times T} \rightarrow \mathbb{R}^{D}$ is a reduction operator that returns a fixed length vector of size $D$ that is independent of $T$. The contextualized embeddings produced from the base model are obtained using the last layer. Other alternatives are also possible, such as ELMo-style re-weighting \citep{peters-arxiv-2018}, through we do not consider these. We consider three types of sentence encoders $\Psi_L$:

\noindent \textbf{Average Pooling.} We consider a parameter-less mean pooling baseline given by $s = \text{meanpool}\{E\}_T$. This baseline is used in all of our experiments as a `uniform' sentence embedding to gauge how a naive sentence encoder performs independent of any context.

\noindent \textbf{Simple (Single Layer).} The simple encoder has only a single parameter matrix $\Theta_L = W \in \mathbb{R}^{D \times K}$. For each timestep $i=1,\ldots,T$, let $F^{(i)} \in \mathbb{R}^D$ denote the transformation of each contextualized word embedding $E^{(i)}$ given by $F^{(i)} =\phi(W E^{(i)} + b)$ where $\phi$ is a non-linearity such as ReLU. This results in a transformed embedding matrix $F \in \mathbb{R}^{D \times T}$. The final $D$-dimensional representation is obtained by max-pooling as $s = \text{maxpool}\{F\}_T$. Note that the use of maxpool has been empirical shown by \citet{conneau-emnlp-2017} and \citet{kiros2018inferlite} to significantly outperform other pooling operations when learning sentence embeddings for downstream tasks, provided there are learnable parameters before the operation.

\noindent \textbf{GatedConv (Self-Attention).} The above sentence encoder treats each contextualized word embedding of $E$ equally. For certain choices of context, this is likely sub-optimal. As a final choice of sentence encoder, we consider a simplified form of the encoder of InferLite \citep{kiros2018inferlite}, which utilizes a fine-grained gating mechanism. Depending on the context, certain features may benefit from either being up-weighted or down-weighted so as to only highlight the relevant aspects of the contextualized embeddings for the given context. The original InferLite encoder take multiple embedding types as input. We consider the special case of a single embedding type. It's possible our method can generalize to multiple types of contextualized embeddings.

This encoder has four components: an encoder, controller, fusion and reduction. In the special case of a single embedding matrix $E$, the encoder and controller take similar forms for $i=0,\ldots,M$ layers:
\begin{eqnarray}
H_0 &=& \phi_{0,h}(W_{0,h} E + b_{0,h}) \\
H_i &=& \phi_{i,h}(W_{i,h} * H_{i-1} + b_{i,h})
\end{eqnarray}
where * denotes a convolution and $H_i$ is the $i$-th layer. The controller has an identical structure to compute $G_0$ and $G_i$. The only difference is the non-linearity on the last layer $G_M$ of the controller is a sigmoid while on the encoder $H_M$ it is tanh. This structure mimics the gated convolutional layers of \citet{van2016conditional}, \citet{dauphin-arxiv-2016} and \citet{gehring-arxiv-2017}. The fusion layer computes a weighted combination plus a skip connection given by
\begin{eqnarray}
F' &=& H_M \odot G_M + G_0 \\
F &=& \phi_f(W_f F' + b_f)
\end{eqnarray}
with the final reduction as $s = \text{maxpool}\{F\}_T$. The GatedConv encoder is essentially a form of self-attention layer.

\subsection{Relatedness lists}

\begin{table*}[t]
  \medskip
  \small
  \centering
  \begin{tabular}{l c c c c c c c c c r}
  \toprule
  Model & MR & CR & SUBJ & MPQA & SST & TREC & MRPC & STSb & SICK & $\Delta$ \\
  \midrule
  BERT CLS \citep{reimers2019sentence} & 78.68 & 84.85 & 94.21 & 88.23 & 84.13 & 91.4 & 71.13 & 16.50 & 42.63 & -5.28 \\
  Glove BOW \citep{conneau-emnlp-2017} & 77.25 & 78.30 & 91.17 & 87.85 & 80.18 & 83.0 & 72.87 & 58.02 & 53.76 & -1.88  \\
  BERT BOW \citep{reimers2019sentence} & 78.66 & 86.25 & 94.37 & 88.66 & 84.40 & 92.8 & 69.45 & 46.35 & 58.40 &  \\
  \midrule
  InferSent \citep{conneau-emnlp-2017} & 81.57 & 86.54 & 92.50 & \bf{90.38} & 84.18 & 88.2 & 75.77 & 68.03 & 65.65 & 3.72  \\
  USE \citep{cer-arxiv-2018} & 80.09 & 85.19 & 93.98 & 86.70 & 86.38 & \bf{93.2} & 70.14 & 74.92 & 76.69 & 5.33  \\
  SBERT \citep{reimers2019sentence} & \bf{84.88} & \bf{90.07} & \bf{94.52} & 90.33 & \bf{90.66} & 87.4 & 75.94 & \bf{79.23} & 73.75 & \bf{7.50} \\
  \midrule
  sCL(BERT-Large; NLI) & 83.40 & 89.32 & 93.49 & 89.16 & 89.35 & 91.8 & \bf{76.75} & 73.75 & 72.08 & 6.64 \\
  \bottomrule
  \end{tabular}
  
  \caption{Comparison of USVs on 9 downstream tasks. The first seven tasks are evaluated by training a logistic regression classifier directly on top of the sentence vectors. Performance is accuracy. The last two tasks report Spearman correlation of unsupervised textual similarity. $\Delta$ indicates the mean improvement over all tasks with respect to BERT BOW. Best results per column are bolded. Results that are not ours are obtained from \citet{reimers2019sentence}.}
  \label{tab:downstream}
\end{table*}

So far we have defined the form of the sentence encoders but still need to define how the parameters are learned given a pre-defined context. We consider a special class of contexts which can be expressed as pairs of relatedness or similarity between lists of sentences. For example, in the context of translation similarity, two sentences are similar under this context if one is a translation of another (parallel). In the case of Natural Language Inference, similarity is given by whether two sentences are entailed, neutral or contradictory. Of course not all contexts are easily defined this way, however these act as a proof of concept for the general method. We consider two forms of training modules from which these relatedness lists can be learned: a Classifier and a Ranker. In both cases, pairs of sentences are encoded by a Siamese network. For the Classifier, we take standard practice and produce a single vector as the concatenation of the two vectors along with their component-wise product and absolute difference. This is then fed to a two-layer neural network with softmax layer to predict a pairwise class label. The corresponding errors are backpropagated through the sentence encoders up to the contextualized embeddings. For the Ranker, we use the margin-based loss of VSE++ \citep{faghri-arxiv-2017} which takes a max over contrastive examples. Contrastive examples are taken from the same minibatch and no special effort is made to mine hard negatives.

\subsection{Combinations}

With the above we can mix and match different contextualized embeddings, sentence encoders and contexts from which to learn the lens parameters. Table ~\ref{tab:usrs} highlights the combinations considered in this work with the corresponding notation used.

\section{Experiments}

\begin{table}
  \medskip
  \small
  \centering
  \begin{tabular}{l r}
  \toprule
  Hyper-parameters & Search space \\
  \midrule
  Batch size & \{128, 256, 512\} \\
  Warmup steps & \{1k, 2k, 4k, 8k, 16k\} \\
  Embedding dropout rate & \{0, 0.1, 0.2\} \\
  Gating layer size & \{128, 256, 512\} \\
  Hidden layer size (Classifier) & \{256, 512, 1024\} \\
  Margin (Ranker) & \{0.1, 0.2, 0.3\} \\
  \bottomrule
  \end{tabular}
  
  \caption{Hyperparameters ranges evaluated by random search.}
  \label{tab:hyper}
\end{table}

We perform experiments across five evaluation settings. Each setting aims to analyze a particular aspect of our models as well as compare to relevant work, namely Sentence BERT \citep{reimers2019sentence} and LASER \citep{artetxe2019massively}.

\subsection{Learning Lensed Sentence Vectors}

The overview of our experimental protocol is as follows. We first train sentence encoders as described in Table ~\ref{tab:usrs}. After training is complete, the composition of contextualized embeddings with their corresponding reduction module is used to map text into fixed-length vectors. These vectors are then directly evaluated on a suite of downstream tasks, similar to other work on universal sentence vectors. No additional fine-tuning is done in any experiments. We use the last layer of publicly available BERT-Large and Multilingual BERT (mBERT) models for contextualized embeddings. Models that lens with NLI are trained on the concatenation of SNLI \citep{bowman-emnlp-2015} and MultiNLI \citep{williams-naacl-2018}. For translation lensing, we follow the instructions of \citet{artetxe2019massively} to reproduce the corpus used for training LASER. This corpus consists of data from Europarl, United Nations Parallel Corpus, OpenSubtitles2018, Global Voices, Tanzil and Tatoeba, all available from OPUS \footnote{\url{http://opus.nlpl.eu/}}. See the appendix of \citet{artetxe2019massively} for complete details. For Tatoeba, we verify that none of our training sentences appear in the official test set. After collecting this corpus, we shuffle the dataset into chunks of size 10k, 100k, 1M, 10M and 100M, representing five orders of magnitude of parallel data. This data is then used for learning the translation-based sentence encoder. We train models with both Simple and GatedConv encoders and dimensionalities of 1024 and 4096. All of our models are trained on a single GPU each using a random hyper-parameter search. Table ~\ref{tab:hyper} describes this search space. All models are trained using Adam \citep{kingma-iclr-2015} with a learning rate warmup schedule \citep{vaswani-nips-2017}. The sentence embedding size determines the initial learning rate. We validate matching errors on a small set of WMT news datasets. For models trained with 100M sentences, we perform just a single pass through the dataset. For NLI-based models, we early stop based on the development set performance. No `peeking' at downstream tasks is done, we exclusively use these validation metrics to choose models for all subsequent experiments.

\subsection{English Downstream Evaluations}

Our first set of experiments compares NLI-lensed sentence embeddings against existing universal sentence vectors. In particular, we focus on comparison with Sentence BERT (SBERT) on a suite of SentEval benchmarks \citep{conneau-arxiv-2018}. The SBERT framework here is essentially identical to ours with the exception that their model fine-tunes the BERT parameters. Thus, we can control other conditions (such as NLI training) and directly analyze what effect fine-tuning all of BERT has on downstream performance vs frozen contextualized embeddings. Other work such as \citet{peters2019tune} perform extensive analysis of `hot' vs `cold' settings but to our knowledge we are the first to explore this for universal sentence vectors. We refer the reader to \citet{reimers2019sentence} for a more detailed description of experimental setup as ours is identical to theirs. 

Results of these experiments are in Table ~\ref{tab:downstream}. Here we make two observations. First, our results outperform on aggregate against existing non-BERT universal sentence encoders while only learning a single weight matrix on top of the contextualized embeddings. Second, while SBERT outperforms on aggregate, the $\Delta$ is small relative to a basic BERT BOW baseline. This demonstrates that the gain from fine-tuning all of BERT for these tasks are minimal and much of the performance improvements over a naive BERT BOW baseline can be achieved by adapting fixed contextualized embeddings. We note that \citet{reimers2019sentence} observed a similar quirk comparing BERT vs RoBERTa \citep{liu2019roberta}. While fine-tuning RoBERTa on static tasks substantially improves over BERT, the gains are minimal for downstream sentence evaluations. Taken together, these results suggest that the fine-tune paradigm that is so crucial for static evaluations may not directly carry over to learning USVs.

\subsection{Matching High-Resource Languages}

The remainder of our experiments are done with multilingual data. For these experiments, we analyze several of our models on matching parallel sentences on 6 languages: English, French, Spanish, German, Czech and Russian. We use WMT newstest2012 as our test corpus which has 6-way parallel data allowing us to experiment with all language pair combinations. For each sentence in the source language, we search for the nearest sentence by cosine similarity. If it is a translation, it is considered a hit. Otherwise it is a miss. We report the average match error.

\begin{figure}[t]
  \centering
  \mbox{
    \subfigure[English]{\includegraphics[width=0.24\textwidth]{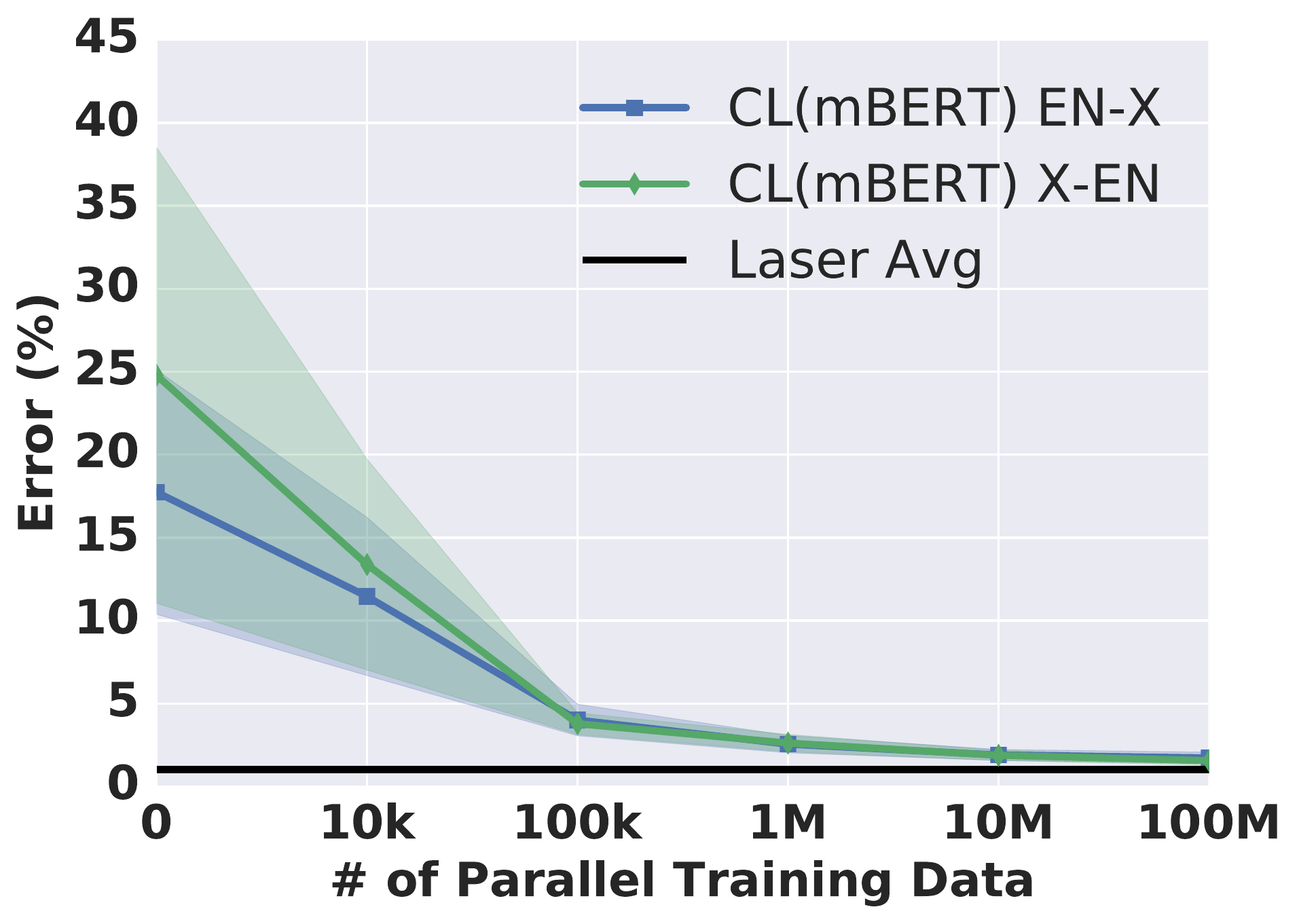}}
    \subfigure[French]{\includegraphics[width=0.24\textwidth]{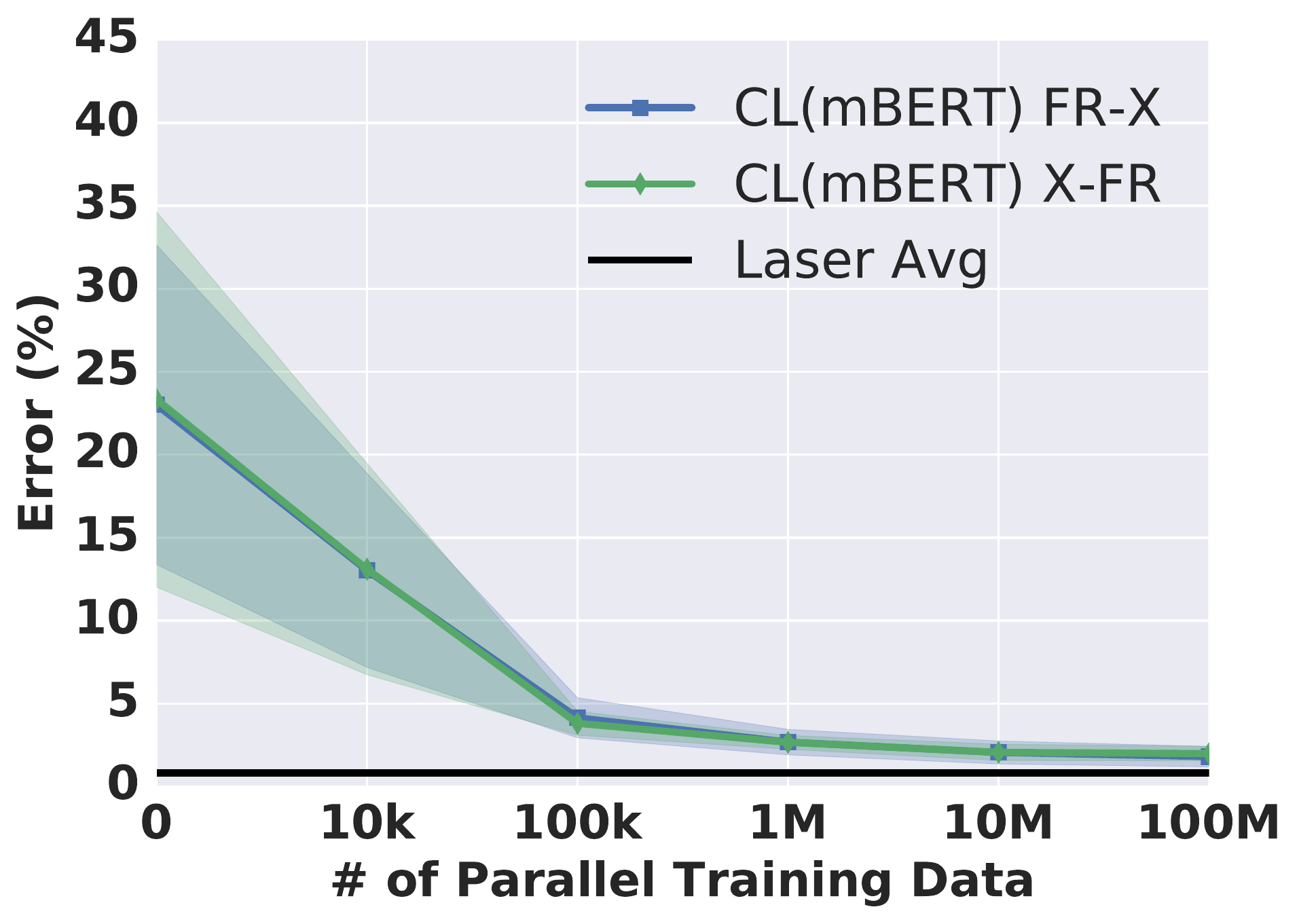}}
  }
  \mbox{
    \subfigure[Spanish]{\includegraphics[width=0.24\textwidth]{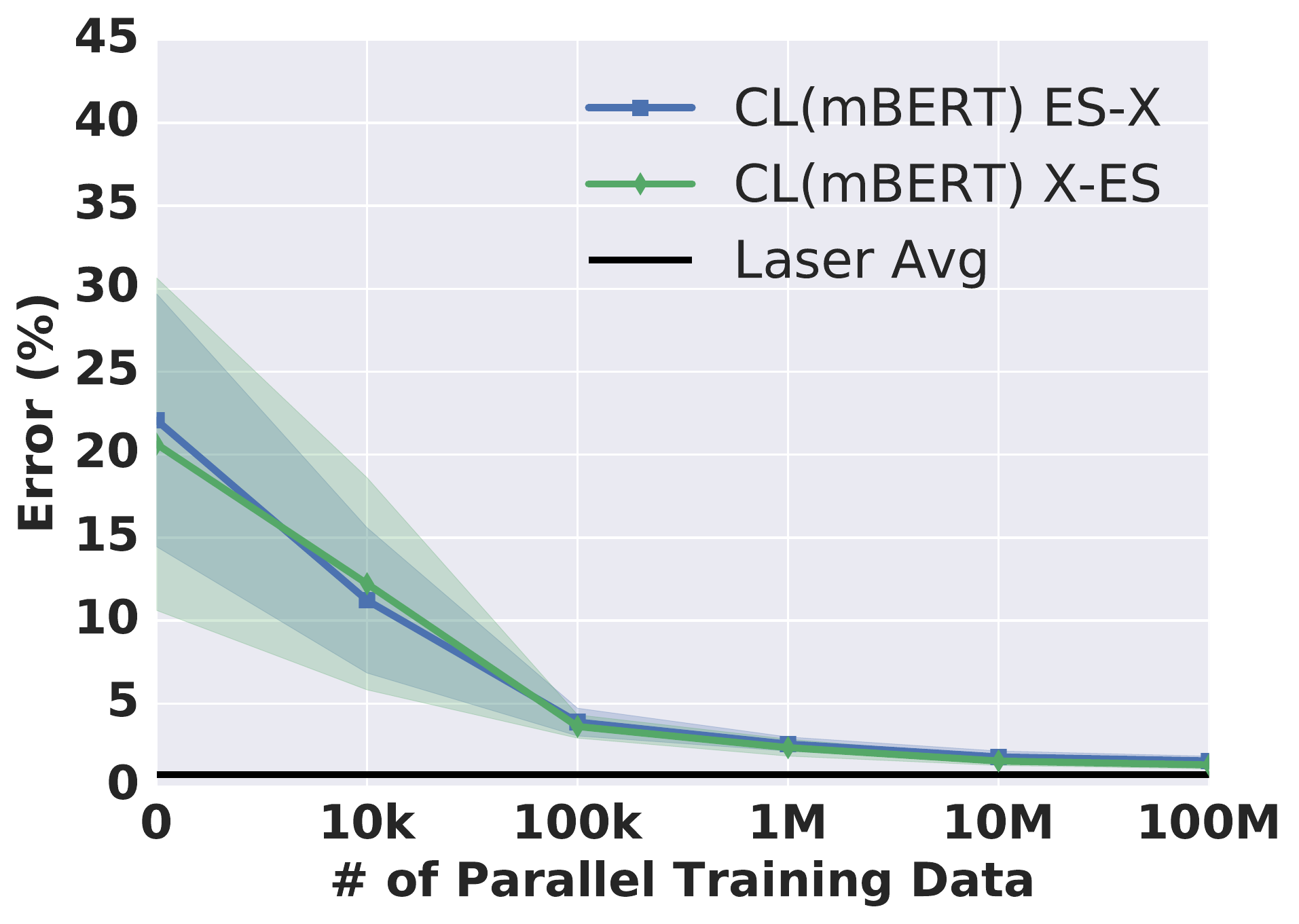}}
    \subfigure[German]{\includegraphics[width=0.24\textwidth]{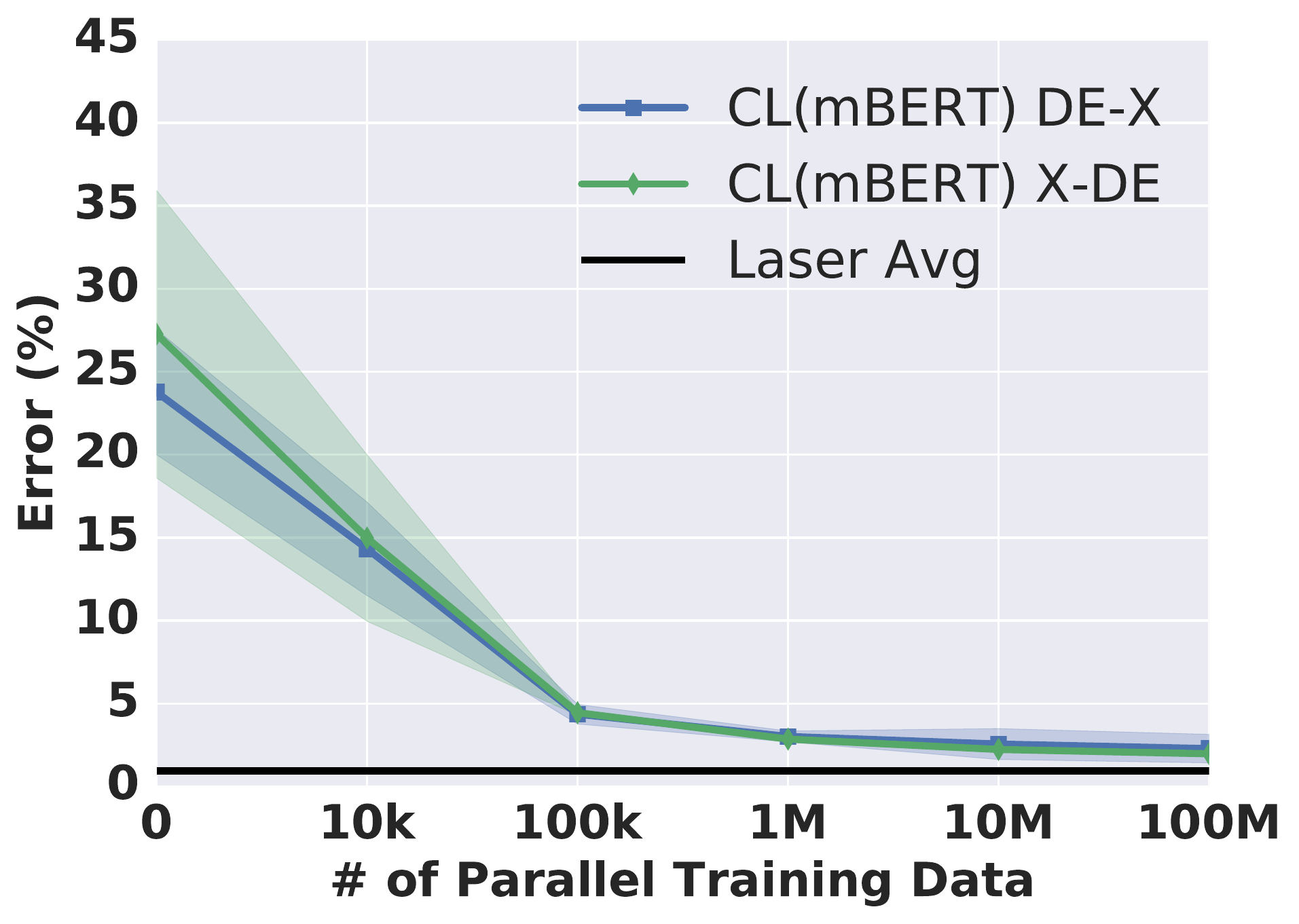}}
  }
  \mbox{
    \subfigure[Czech]{\includegraphics[width=0.24\textwidth]{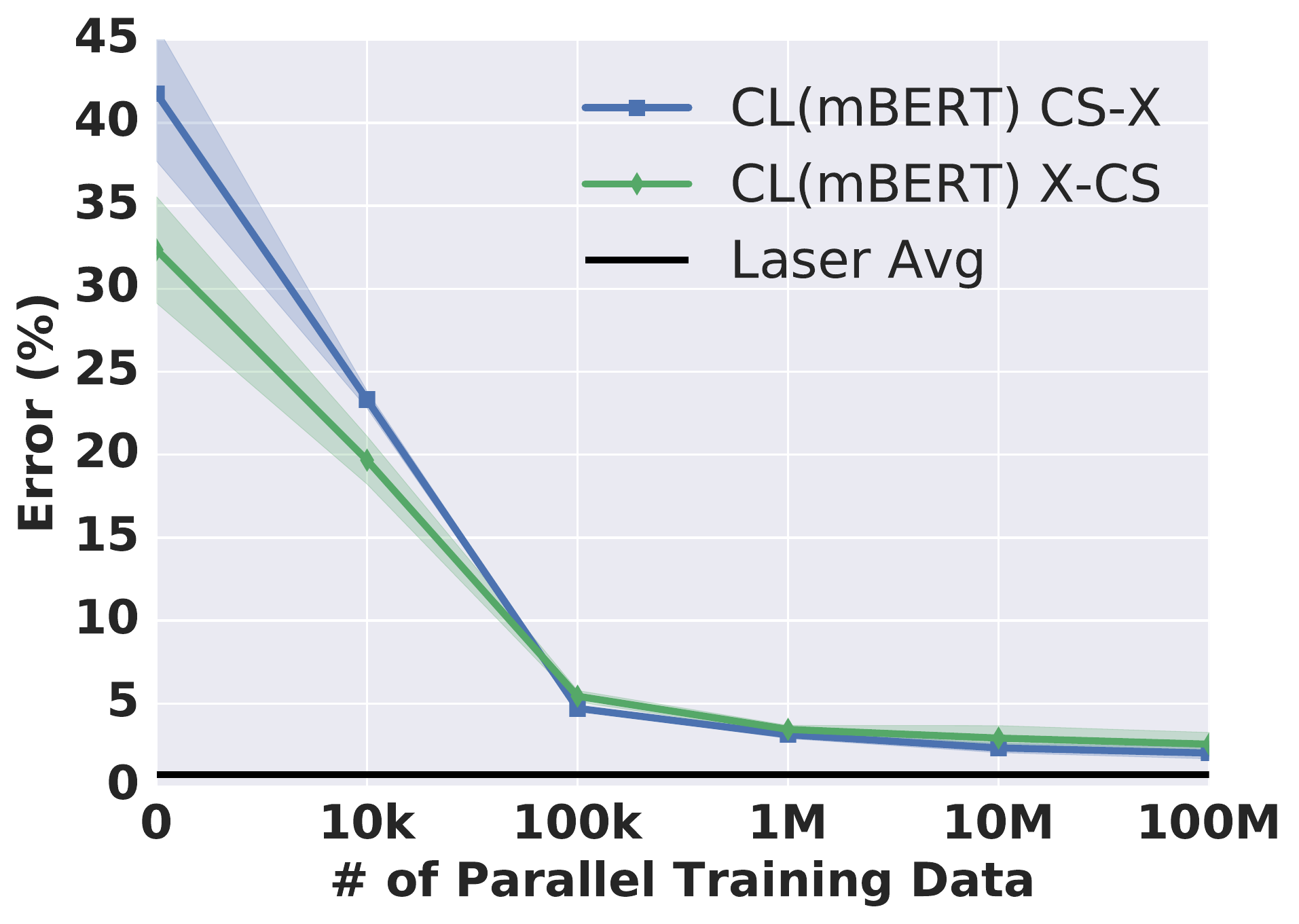}}
    \subfigure[Russian]{\includegraphics[width=0.24\textwidth]{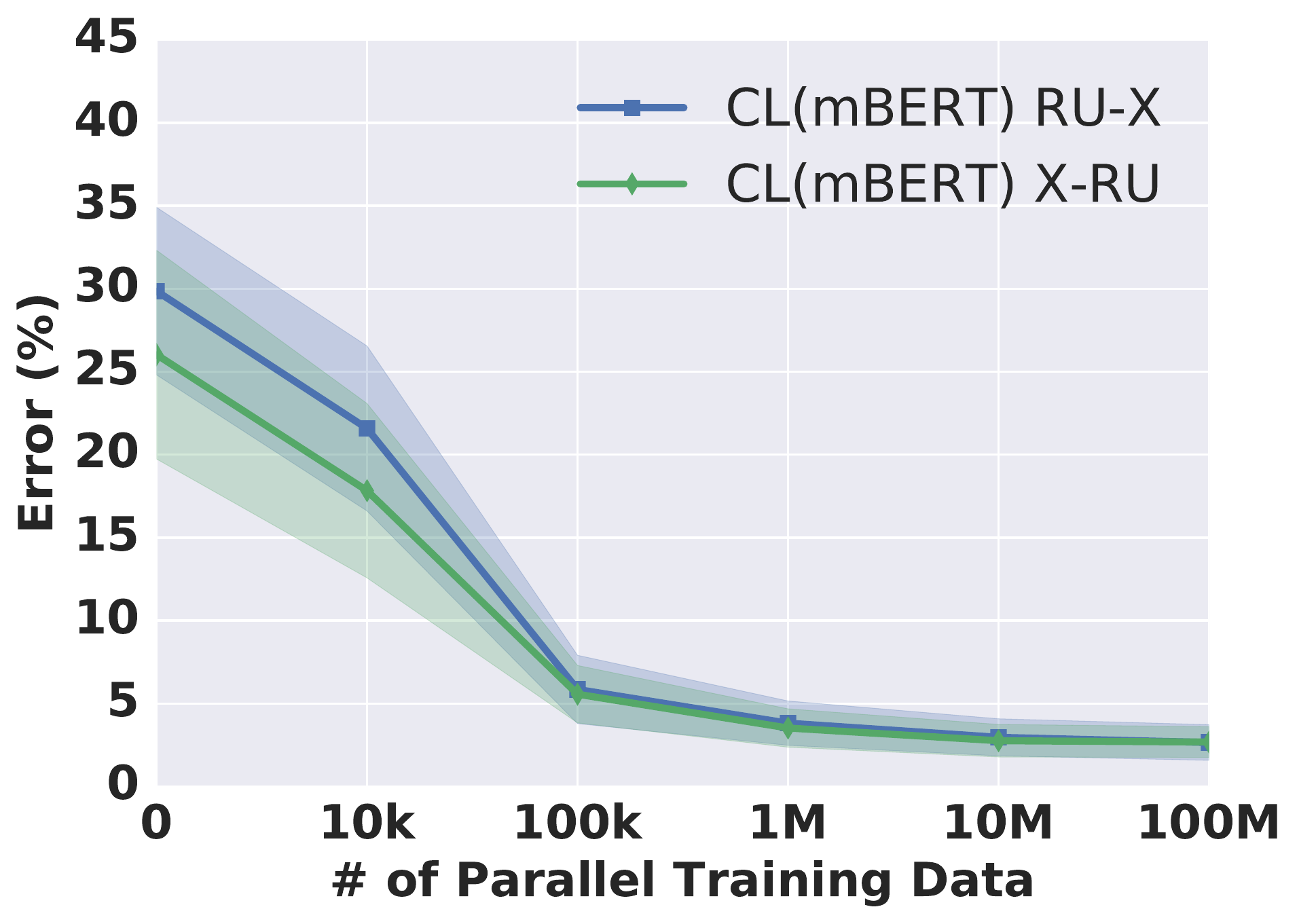}}
  }
  \caption{Error rates from searching for the correct ground-truth translation across 6 languages for WMT newstest2012. Standard deviation is computed over all other languages from the source/target. Averaged error rates from LASER are included. }
\label{fig:wmt}
\end{figure}

Figure ~\ref{fig:wmt} shows results on each language using 6 of our sentence encoders with varying amounts of parallel data. We also plot LASER performance. Observe that as we increase the amount of parallel data, performance improves and converges to within 0.5-1.0 absolute error to LASER. Also observe that unsupervised encoders (no parallel data) are able to achieve mean errors between 20-40\%. While substantially worse than parallel models, these results add to a growing body of work the implicit translation abilities of large monolingual self-supervision \citep{pires2019multilingual}.

\begin{figure*}[t]
  \centering
  \mbox{
    \subfigure[Avg - NLI]{\includegraphics[width=0.24\textwidth]{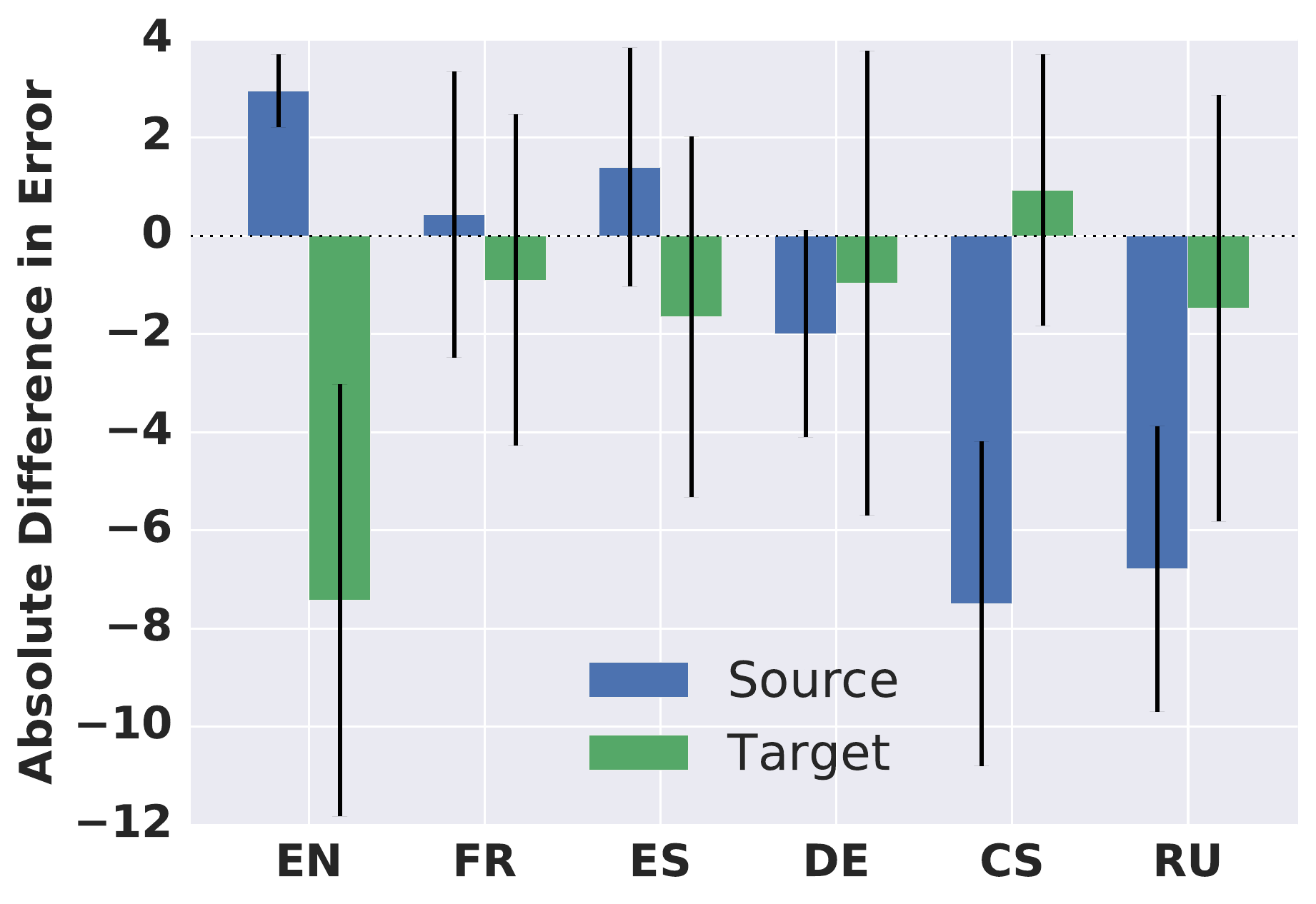}}
    \subfigure[GatedConv - Simple]{\includegraphics[width=0.24\textwidth]{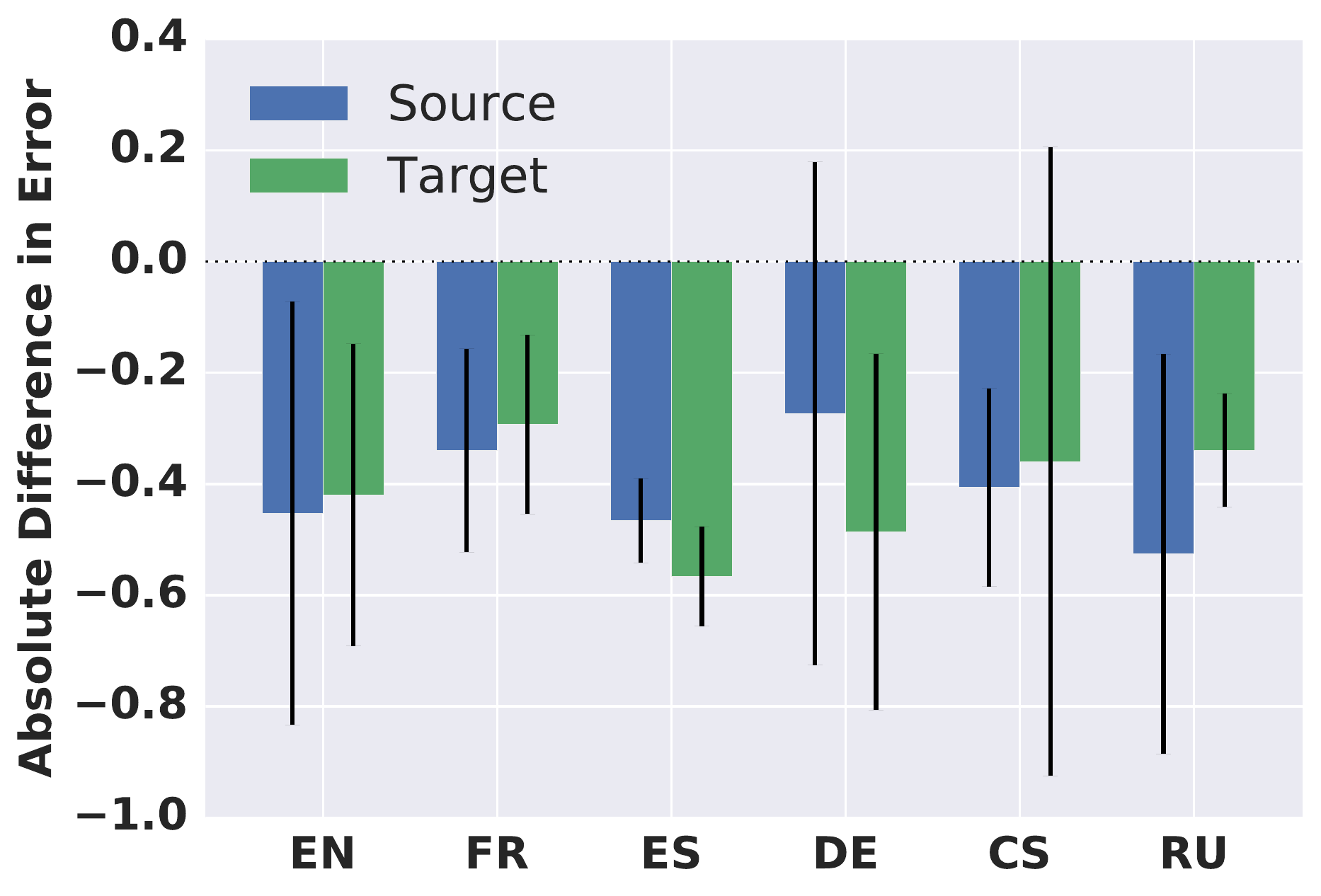}}
    \subfigure[1024 dim - 4096 dim]{\includegraphics[width=0.24\textwidth]{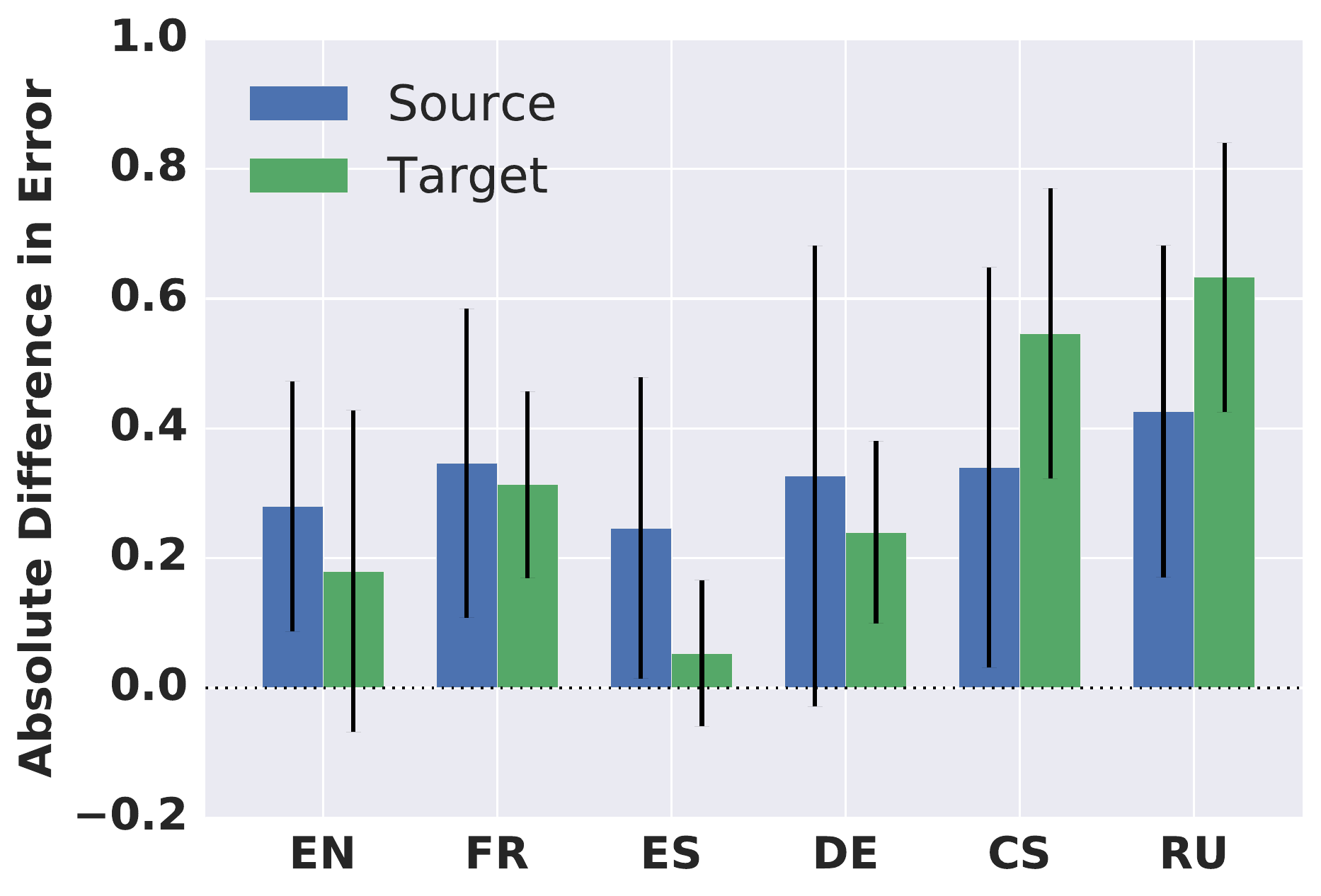}}
    \subfigure[Dense - Binary]{\includegraphics[width=0.24\textwidth]{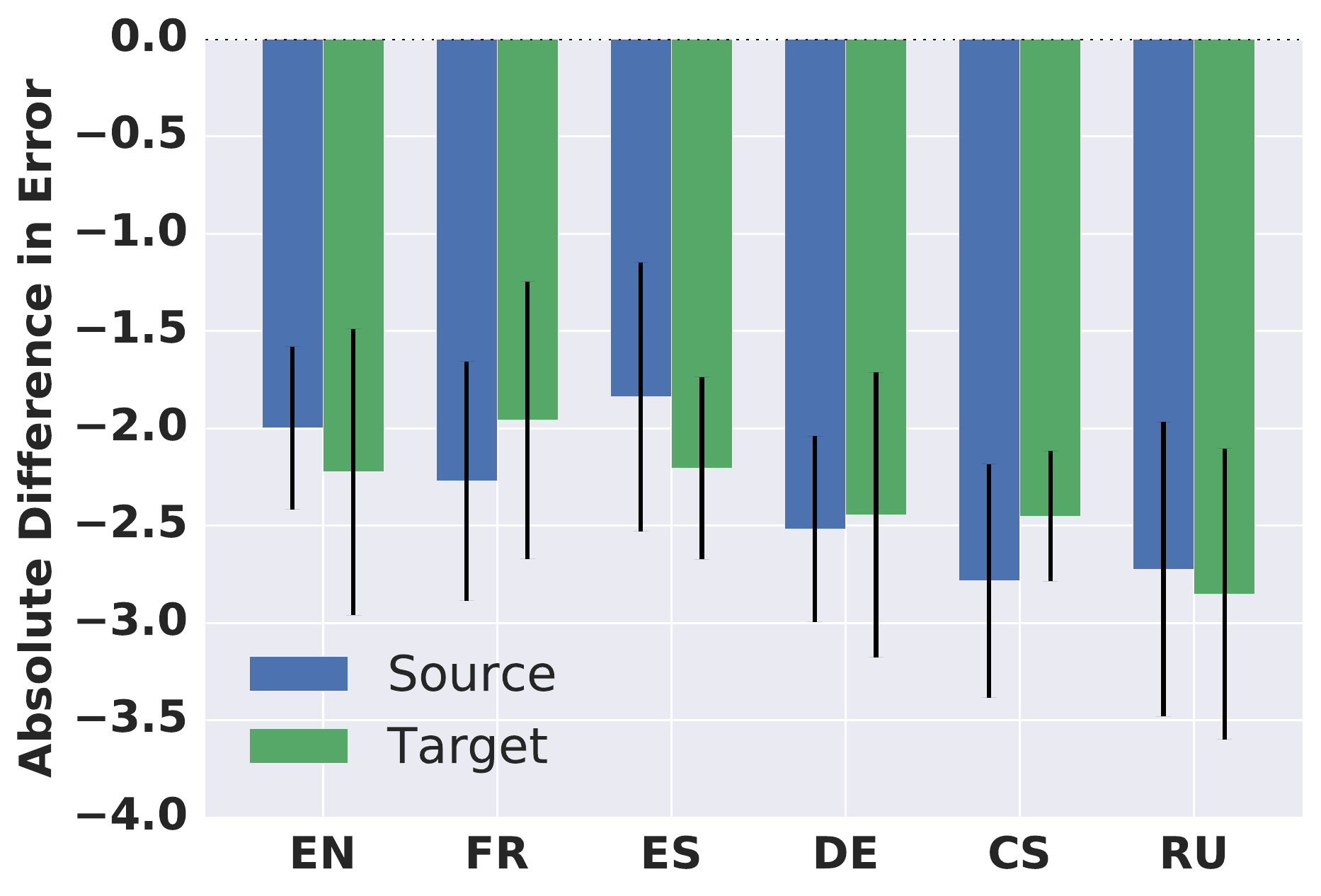}}
}
  \caption{Relative comparisons of different model types. (a): mBERT mean pooled embeddings against NLI (b) GatedConv encoder against simple encoder (c) 1024-dimensional vectors against 4096-dimensional vectors and (d) dense vectors vs sparse binary vectors. Positive numbers indicate the right argument performs better while negative numbers indicate the left argument performs better.}
\label{fig:wmtcompare}
\end{figure*}

We also do relative comparisons to different components of our models. In Figure ~\ref{fig:wmtcompare} we analyze 4 comparisons: unsupervised models (BERT BOW vs NLI), encoder type (GatedConv vs Simple), sentence embedding dimensionality (1024 vs 4096) and embedding type (Dense vs Sparse binary). Binary representations are obtained by thesholding at 1.0, resulting in a vector with 2.5\% active units. This works due to the ReLU+MaxPool combination of the sentence encoder. See \citet{shen2019learning} for an extensive analysis of binary sentence embeddings.

\subsection{Mining Parallel Sentences}

\begin{table}[t]
  \medskip
  \small
  \centering
  \begin{tabular}{l c c c}
  \toprule
  Language & mBERT? & LASER & Ours \\
  \midrule
  Algerian Arabic & & \textbf{60.5} & 71.5 \\
  Asturian & \checkmark & 13.8 & \textbf{7.5} \\
  Awadhi & & 63.9 & \textbf{62.1} \\
  Cebuano & \checkmark & 84.3 & \textbf{81.8} \\
  Chamorro & & 70.8 & \textbf{65.7} \\
  Egyptian Arabic & & \textbf{31.1} & 45.5 \\
  Faroese & & 28.4 & \textbf{25.4} \\
  Gaelic; Scottish Gaelic & & 96.3 & \textbf{84.3} \\
  Javanese & \checkmark & 77.1 & \textbf{49.5} \\
  Kashubian & & 56.7 & \textbf{48.8} \\
  Mongolian & \checkmark & 91.8 & \textbf{47.7} \\
  North Moluccan Malay & & 49.1 & \textbf{44.7} \\
  Novial & & 34.1 & \textbf{25.1} \\
  Nynorsk Norwegian & \checkmark & 11.7 & \textbf{8.3} \\
  Old English & & 62.3 & \textbf{39.9} \\
  Pampangan; Kapampangan & & 94.1 & \textbf{90.1} \\
  Piemontese & & 50.4 & \textbf{22.1} \\
  Russian old & & 71.9 & \textbf{64.7} \\
  Sorbian Lower & & 52.0 & \textbf{42.3} \\
  Sorbian Upper & & 45.5 & \textbf{36.5} \\
  Swabian & & 54.0 & \textbf{34.8} \\
  Swiss German & & \textbf{55.6} & 57.7 \\
  Talossan & & 55.3 & \textbf{44.2} \\
  Turkmen & & 79.3 & \textbf{69.2} \\
  Waray & \checkmark & 86.4 & \textbf{79.5} \\
  Welsh & \checkmark & 91.4 & \textbf{52.9} \\
  Western Frisian & \checkmark & 48.3 & \textbf{16.8} \\
  Xhosa & & 91.6 & \textbf{84.2} \\
  Yiddish & & \textbf{94.3} & 94.5 \\
  \bottomrule
  \end{tabular}
  
  \caption{Match errors for languages from which there is no parallel data. For each language, we also indicate whether the language was included for self-supervision as part of multilingual BERT.}
  \label{tab:noparallel}
\end{table}

\begin{table*}[t]
  \medskip
  \small
  \centering
  \begin{tabular}{l c c c c c c c c r r}
  \toprule
  Model & DE & FR & RU & ZH & DE & FR & RU & ZH & Mean($\tau$) & Std($\tau$) \\
  \midrule
  \citet{schwenk2018filtering} & 76.1 & 74.9 & 73.3 & 71.6 & 76.9 & 75.8 & 73.8 & 71.6 & & \\
  \citet{azpeitia2018extracting} & 84.27 & 80.63 & 80.89 & 76.45 & 85.52 & 81.47 & 81.30 & 77.45 & & \\
  LASER \citep{artetxe2019massively} & \bf{95.43} & \bf{92.40} & \bf{92.29} & \bf{91.20} & \bf{96.19} & \bf{93.91} & \bf{93.30} & \bf{92.27} & & \\
  \midrule
  mBERT BOW & 54.47 & 50.89 & 39.42 & 39.57 & & & & & 1.11 & 0.012 \\
  CL(mBERT; NLI) & \underline{59.00} & \underline{59.46} & \underline{47.11} & \underline{41.12} & & & & & 1.13 & 0.012 \\
  \midrule
  CL(mBERT; 100M) & 88.24 & 85.11 & 86.18 & 82.15 & & & & & 1.103 & 0.004 \\
  sCL(mBERT; 100M)$\dagger$ & \underline{89.98} & 86.79 & 87.15 & 84.91 & & & & & 1.130 & 0.007 \\
  CL(mBERT; 100M)$\dagger$ & 89.84 & \underline{87.14} & \underline{88.00} & \underline{86.05} & 90.24 & 88.54 & 89.25 & 86.70 & 1.132 & 0.011 \\
  \bottomrule
  \end{tabular}
  
  \caption{F1 scores for parallel data mining on the BUCC 2018 training and test sets. The first four language columns are training while the last four are testing. Middle group are unsupervised mining. For training data we report oracle scores with the corresponding mean and variance of thresholds ($\tau$). For test data, we report results using the best threshold found on the training sets. Best results overall are bolded, best results per group are underlined.}
  \label{tab:bucctrain}
\end{table*}

In this experiment, we consider the task of parallel sentence mining. Given two monolingual corpora in different languages, the goal is to return (hard mine) pairs of sentences which are direct translations of each other while ignoring all other pairs. We use the BUCC 2018 corpora \citep{zweigenbaum2018overview} for this task, consisting of a target English corpus and source data in German, French, Russian and Chinese. The metric is F1 score. A naive way to perform this task is just to compute all-pairs cosine similarity and then choose a threshold. However this often performs poorly due to the hubness problem. Instead, we use a type of normalized similarity that mitigates the hubness problem by re-weighting scores based on scores of its nearest neighbours. To do this, we use margin-based scoring of \citet{artetxe2018margin}:
\begin{multline*}
    \score(x, y) = \margin (\cos(x, y), \\
    \sum_{z \in \nn_k(x)}{\frac{\cos(x, z)}{2k}} +  \sum_{z \in \nn_k(y)}{\frac{\cos(y, z)}{2k}})
\end{multline*}
with \textit{ratio} ($\margin(a, b) = \frac{a}{b}$) score, a generalization of CSLS proposed by \citet{conneau2017word}. Here, $x$ and $y$ are two sentences and $\nn_k(x)$ are the $k$-nearest neighbours of $x$. We use $k=4$ in all experiments. For unsupervised models, we report `oracle' results using the best threshold on the training set. This is to our knowledge the first attempt at unsupervised parallel sentence mining.

Our results are reported in Table ~\ref{tab:bucctrain}. First observe the mean and variance of the threshold parameters. They are nearly identical across all languages, a similar result observed by \citet{artetxe2019massively}. This is even true for unsupervised models, meaning one could tune the threshold on high-resource pairs and use the same model for mining language pairs without seeing any parallel data. Again, we see strong performance from our Simple encoder, which only learns a single weight matrix. In this experiment, we notice a more substantial performance difference between our best results and LASER. We made a single submission of our best model to \citet{zweigenbaum2018overview} to be evaluated on a held-out test set.

\subsection{Matching on 100+ Languages}

\begin{figure*}[t]
  \centering
  \mbox{
    \subfigure{\includegraphics[width=0.99\textwidth]{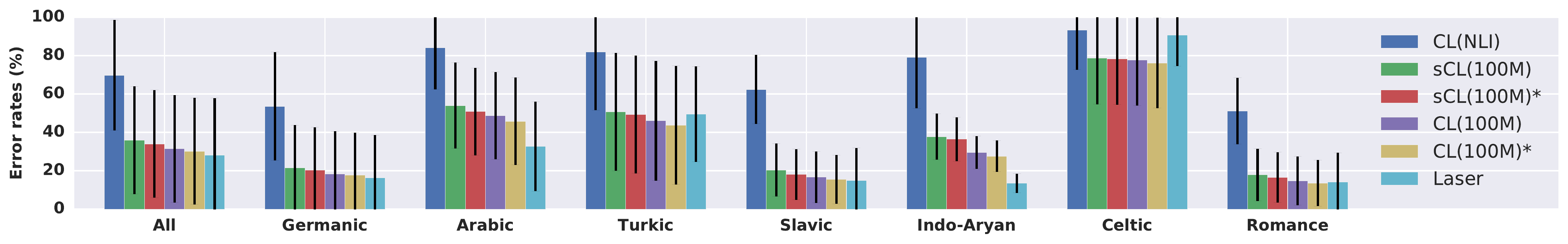}}
}
  \mbox{
    \subfigure{\includegraphics[width=0.99\textwidth]{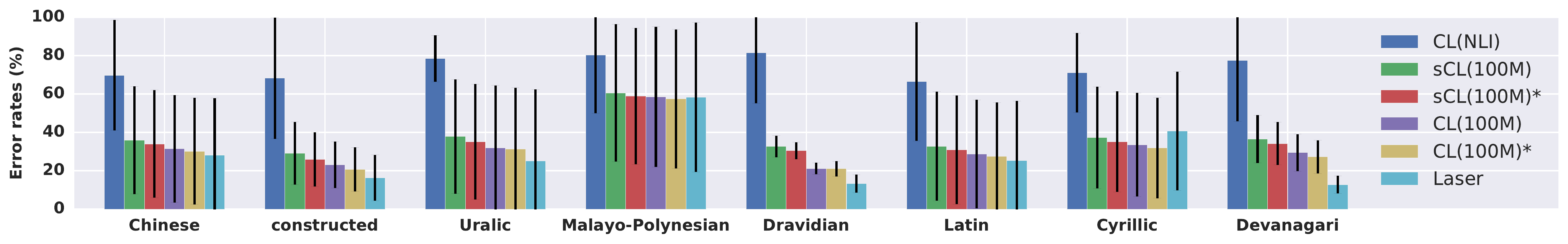}}
}
  \caption{Match errors for different models across language groupings on Tatoeba. All indicates all languages. The last three groups are scripts. All others are language families. A * next to the method indicates that it uses a 4096-dimensional vector.}
\label{fig:tatbar}
\end{figure*}

As of now we have only explored experiments on high resource languages. We next consider matching experiments on the Tatoeba test dataset, consisting of 112 languages. The evaluation is identical to before: we take cosine similarities across language pairs and compute mean matching errors based on whether we retrieve the correct sentence translation. In Figure ~\ref{fig:tatbar} we report results as a function of language families and scripts for a number of our models compared to LASER. In Table ~\ref{tab:noparallel} we consider languages from which there is no parallel data. 

Across all languages, we obtain a similar overall match error to LASER. On some families LASER is better while on others ours is. Perhaps most interestingly are results across languages with no parallel data. Here we observe substantial gains over LASER, noting which languages were also available to mBERT. Taken together, we observe that (a) LASER is better for high-resource languages (b) mixed across certain language families and (c) ours is better for very low resource languages. The later highlights the usefulness of contextualized embeddings vs learning all parameters of the sentence encoder directly from parallel data.

\subsection{Visualization of Language Embeddings}

\begin{figure*}[t]
  \centering
  \mbox{
    \subfigure{\includegraphics[width=0.99\textwidth]{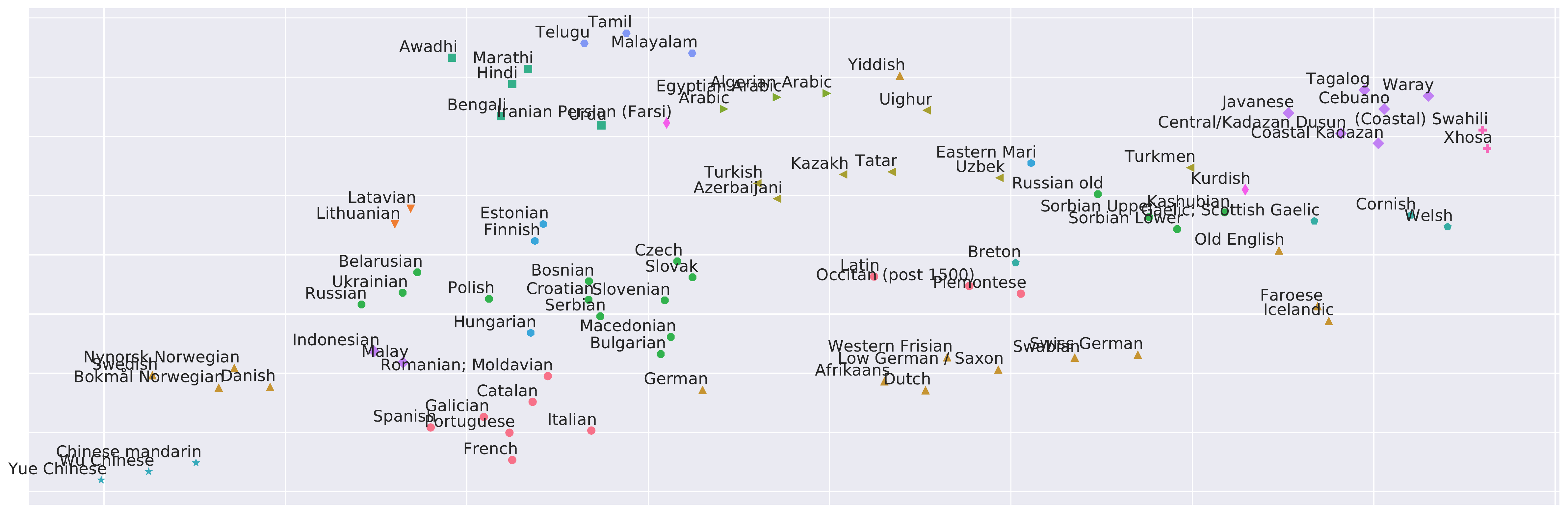}}
}
  \caption{t-SNE of language vectors colored by language family. No parallel data is used for these embeddings. Best seen electronically.}
\label{fig:tsne}
\end{figure*}

As a final experiment, we qualitatively analyze unsupervised language embeddings derived from Tatoeba test sentences. For test sentences in each language, we first normalize each sentence embedding to have unit length, then compute the variance across all sentence embeddings in the corresponding language. This is to derive an embedding that is independent of the particular sentences from that language. We then apply t-SNE \citep{maaten2008visualizing} to obtain a two-dimensional visualization of these language embeddings. This is done using the CL(mBERT; NLI) model which has no access to parallel data. The visualization is in Figure ~\ref{fig:tsne}. Each tick is color coded to its language family. This is for visualization only, the model has no access to this information. We observe a surprising amount of family relatedness, indicating that our embeddings can capture high-level properties of the underlying language families. A similar visualization is also done by \citep{kudugunta2019investigating} under different settings. We note that we also created a similar visualization with models that had access to parallel data and observed similar outputs. This suggests that high-level language relatedness was primarily learned through self-supervision with minimal effects from the addition of parallel data.

Finally, we took the Tatoeba test sentences and trained a logistic regression classifier to predict what language each vector encodes. Using 1\% of sentences to train, we observed that mBERT BOW could obtain over 50\% accuracy, CL(mBERT; NLI) achieves 23\% accuracy while CL(mBERT; 100M) was only 10\% accuracy. Models using the Simple encoder had a higher accuracy than those with a GatedConv encoder. Our results suggest that translation-based lensing removes language discrimination to some degree but largely keeps languages in separate subspaces.

\section{Conclusion}

Our work proposed a framework for obtaining lensed universal sentence vectors as a function of a pre-defined context. Experimentally we show strong results against existing methods in a wide range of settings. As highlighted in the introduction, our work is not strictly about translation or NLI. These are test beds from which we can perform experiments relative to existing methods. Our framework can be adapted to most notions of context. Future work would aim to study dynamic lensing and other settings such as few-shot learning and non-explicit contextualization. It would also be of interest to compare other contextualized embedding models that have substantially outperformed mBERT \citep{conneau2019cross, conneau2019unsupervised}. Using these as base models could result in even higher quality universal sentence vectors.

\section*{Acknowledgements}

The author would like to thank Geoff Hinton, Mohammad Norouzi and Felix Hill for their feedback.

\bibliography{example_paper}

\begin{thebibliography}{54}
\providecommand{\natexlab}[1]{#1}
\providecommand{\url}[1]{\texttt{#1}}
\expandafter\ifx\csname urlstyle\endcsname\relax
  \providecommand{\doi}[1]{doi: #1}\else
  \providecommand{\doi}{doi: \begingroup \urlstyle{rm}\Url}\fi

\bibitem[Arora et~al.(2017)Arora, Liang, and Ma]{arora-iclr-2017}
Arora, S., Liang, Y., and Ma, T.
\newblock {A Simple but Tough-to-Beat Baseline for Sentence Embeddings}.
\newblock In \emph{{ICLR}}, 2017.

\bibitem[Artetxe \& Schwenk(2018)Artetxe and Schwenk]{artetxe2018margin}
Artetxe, M. and Schwenk, H.
\newblock Margin-based parallel corpus mining with multilingual sentence
  embeddings.
\newblock In \emph{arXiv:1811.01136}, 2018.

\bibitem[Artetxe \& Schwenk(2019)Artetxe and Schwenk]{artetxe2019massively}
Artetxe, M. and Schwenk, H.
\newblock Massively multilingual sentence embeddings for zero-shot
  cross-lingual transfer and beyond.
\newblock In \emph{TACL}, 2019.

\bibitem[Azpeitia et~al.(2018)Azpeitia, Etchegoyhen, and
  Garcia]{azpeitia2018extracting}
Azpeitia, A., Etchegoyhen, T., and Garcia, E.~M.
\newblock Extracting parallel sentences from comparable corpora with stacc
  variants.
\newblock In \emph{11th Workshop on Building and Using Comparable Corpora},
  2018.

\bibitem[Bansal et~al.(2019)Bansal, Jha, and McCallum]{bansal2019learning}
Bansal, T., Jha, R., and McCallum, A.
\newblock Learning to few-shot learn across diverse natural language
  classification tasks.
\newblock In \emph{arXiv:1911.03863}, 2019.

\bibitem[Barsalou(1983)]{barsalou1983ad}
Barsalou, L.~W.
\newblock Ad hoc categories.
\newblock In \emph{Memory \& cognition}, 1983.

\bibitem[Bowman et~al.(2015)Bowman, Angeli, Potts, and
  Manning]{bowman-emnlp-2015}
Bowman, S., Angeli, G., Potts, C., and Manning, C.
\newblock {A large annotated corpus for learning natural language inference}.
\newblock In \emph{{EMNLP}}, 2015.

\bibitem[Cer et~al.(2018)Cer, Yang, Kong, Hua, Limtiaco, John, Constant,
  Guajardo-Cespedes, Yuan, Tar, et~al.]{cer-arxiv-2018}
Cer, D., Yang, Y., Kong, S.-y., Hua, N., Limtiaco, N., John, R.~S., Constant,
  N., Guajardo-Cespedes, M., Yuan, S., Tar, C., et~al.
\newblock {Universal Sentence Encoder}.
\newblock In \emph{arXiv:1803.11175}, 2018.

\bibitem[Chan et~al.(2019{\natexlab{a}})Chan, Kiros, and Chan]{chan2019mkermit}
Chan, H., Kiros, J., and Chan, W.
\newblock Multilingual kermit: It’s not easy being generative.
\newblock In \emph{The 3rd Workshop on Neural Generation and Translation},
  2019{\natexlab{a}}.

\bibitem[Chan et~al.(2019{\natexlab{b}})Chan, Kitaev, Guu, Stern, and
  Uszkoreit]{chan2019kermit}
Chan, W., Kitaev, N., Guu, K., Stern, M., and Uszkoreit, J.
\newblock Kermit: Generative insertion-based modeling for sequences.
\newblock In \emph{arXiv:1906.01604}, 2019{\natexlab{b}}.

\bibitem[Conneau \& Kiela(2018)Conneau and Kiela]{conneau-arxiv-2018}
Conneau, A. and Kiela, D.
\newblock {SentEval: An Evaluation Toolkit for Universal Sentence
  Representations}.
\newblock In \emph{arXiv:1803.05449}, 2018.

\bibitem[Conneau \& Lample(2019)Conneau and Lample]{conneau2019cross}
Conneau, A. and Lample, G.
\newblock Cross-lingual language model pretraining.
\newblock In \emph{NeurIPS}, 2019.

\bibitem[Conneau et~al.(2017{\natexlab{a}})Conneau, Kiela, Schwenk, Barrault,
  and Bordes]{conneau-emnlp-2017}
Conneau, A., Kiela, D., Schwenk, H., Barrault, L., and Bordes, A.
\newblock {Supervised Learning of Universal Sentence Representations from
  Natural Language Inference Data}.
\newblock In \emph{{EMNLP}}, 2017{\natexlab{a}}.

\bibitem[Conneau et~al.(2017{\natexlab{b}})Conneau, Lample, Ranzato, Denoyer,
  and J{\'e}gou]{conneau2017word}
Conneau, A., Lample, G., Ranzato, M., Denoyer, L., and J{\'e}gou, H.
\newblock Word translation without parallel data.
\newblock In \emph{arXiv:1710.04087}, 2017{\natexlab{b}}.

\bibitem[Conneau et~al.(2019)Conneau, Khandelwal, Goyal, Chaudhary, Wenzek,
  Guzm{\'a}n, Grave, Ott, Zettlemoyer, and Stoyanov]{conneau2019unsupervised}
Conneau, A., Khandelwal, K., Goyal, N., Chaudhary, V., Wenzek, G., Guzm{\'a}n,
  F., Grave, E., Ott, M., Zettlemoyer, L., and Stoyanov, V.
\newblock Unsupervised cross-lingual representation learning at scale.
\newblock In \emph{arXiv:1911.02116}, 2019.

\bibitem[Dai \& Le(2015)Dai and Le]{dai2015semi}
Dai, A.~M. and Le, Q.~V.
\newblock {Semi-supervised sequence learning}.
\newblock In \emph{NIPS}, 2015.

\bibitem[Dauphin et~al.(2016)Dauphin, Fan, Auli, and
  Grangier]{dauphin-arxiv-2016}
Dauphin, Y.~N., Fan, A., Auli, M., and Grangier, D.
\newblock {Language modeling with gated convolutional networks}.
\newblock In \emph{arXiv:1612.08083}, 2016.

\bibitem[Devlin et~al.(2018)Devlin, Chang, Lee, and Toutanova]{devlin2018bert}
Devlin, J., Chang, M.-W., Lee, K., and Toutanova, K.
\newblock Bert: Pre-training of deep bidirectional transformers for language
  understanding.
\newblock In \emph{arXiv:1810.04805}, 2018.

\bibitem[Faghri et~al.(2017)Faghri, Fleet, Kiros, and
  Fidler]{faghri-arxiv-2017}
Faghri, F., Fleet, D., Kiros, J., and Fidler, S.
\newblock {VSE++: Improving Visual-Semantic Embeddings with Hard Negatives}.
\newblock In \emph{arXiv:1707.05612}, 2017.

\bibitem[Gehring et~al.(2017)Gehring, Auli, Grangier, Yarats, and
  Dauphin]{gehring-arxiv-2017}
Gehring, J., Auli, M., Grangier, D., Yarats, D., and Dauphin, Y.~N.
\newblock {Convolutional sequence to sequence learning}.
\newblock In \emph{arXiv:1705.03122}, 2017.

\bibitem[Guo et~al.(2018)Guo, Shen, Yang, Ge, Cer, Abrego, Stevens, Constant,
  Sung, Strope, et~al.]{guo2018effective}
Guo, M., Shen, Q., Yang, Y., Ge, H., Cer, D., Abrego, G.~H., Stevens, K.,
  Constant, N., Sung, Y.-H., Strope, B., et~al.
\newblock Effective parallel corpus mining using bilingual sentence embeddings.
\newblock In \emph{arXiv:1807.11906}, 2018.

\bibitem[Ha et~al.(2016)Ha, Dai, and Le]{ha2016hypernetworks}
Ha, D., Dai, A., and Le, Q.~V.
\newblock Hypernetworks.
\newblock In \emph{arXiv:1609.09106}, 2016.

\bibitem[Hill et~al.(2016)Hill, Cho, and Korhonen]{hill-naacl-2016}
Hill, F., Cho, K., and Korhonen, A.
\newblock Learning distributed representations of sentences from unlabelled
  data.
\newblock In \emph{{NAACL}}, 2016.

\bibitem[Houlsby et~al.(2019)Houlsby, Giurgiu, Jastrzebski, Morrone,
  De~Laroussilhe, Gesmundo, Attariyan, and Gelly]{houlsby2019parameter}
Houlsby, N., Giurgiu, A., Jastrzebski, S., Morrone, B., De~Laroussilhe, Q.,
  Gesmundo, A., Attariyan, M., and Gelly, S.
\newblock Parameter-efficient transfer learning for nlp.
\newblock In \emph{arXiv:1902.00751}, 2019.

\bibitem[Howard \& Ruder(2018)Howard and Ruder]{howard2018universal}
Howard, J. and Ruder, S.
\newblock {Universal language model fine-tuning for text classification}.
\newblock In \emph{ACL}, 2018.

\bibitem[Kingma \& Ba(2015)Kingma and Ba]{kingma-iclr-2015}
Kingma, D. and Ba, J.
\newblock {Adam: A Method for Stochastic Optimization}.
\newblock In \emph{{ICLR}}, 2015.

\bibitem[Kiros \& Chan(2018)Kiros and Chan]{kiros2018inferlite}
Kiros, J. and Chan, W.
\newblock Inferlite: Simple universal sentence representations from natural
  language inference data.
\newblock In \emph{EMNLP}, 2018.

\bibitem[Kiros et~al.(2015)Kiros, Zhu, Salakhutdinov, Zemel, Torralba, Urtasun,
  and Fidler]{kiros-nips-2015}
Kiros, R., Zhu, Y., Salakhutdinov, R., Zemel, R.~S., Torralba, A., Urtasun, R.,
  and Fidler, S.
\newblock {Skip-Thought Vectors}.
\newblock In \emph{{NIPS}}, 2015.

\bibitem[Kudugunta et~al.(2019)Kudugunta, Bapna, Caswell, Arivazhagan, and
  Firat]{kudugunta2019investigating}
Kudugunta, S.~R., Bapna, A., Caswell, I., Arivazhagan, N., and Firat, O.
\newblock Investigating multilingual nmt representations at scale.
\newblock In \emph{arXiv preprint arXiv:1909.02197}, 2019.

\bibitem[Liu et~al.(2019)Liu, Ott, Goyal, Du, Joshi, Chen, Levy, Lewis,
  Zettlemoyer, and Stoyanov]{liu2019roberta}
Liu, Y., Ott, M., Goyal, N., Du, J., Joshi, M., Chen, D., Levy, O., Lewis, M.,
  Zettlemoyer, L., and Stoyanov, V.
\newblock Roberta: A robustly optimized bert pretraining approach.
\newblock In \emph{arXiv:1907.11692}, 2019.

\bibitem[Logeswaran \& Lee(2018)Logeswaran and Lee]{logeswaran-iclr-2018}
Logeswaran, L. and Lee, H.
\newblock An efficient framework for learning sentence representations.
\newblock In \emph{{ICLR}}, 2018.

\bibitem[Lu et~al.(2019)Lu, Batra, Parikh, and Lee]{lu2019vilbert}
Lu, J., Batra, D., Parikh, D., and Lee, S.
\newblock Vilbert: Pretraining task-agnostic visiolinguistic representations
  for vision-and-language tasks.
\newblock In \emph{NeurIPS}, 2019.

\bibitem[Maaten \& Hinton(2008)Maaten and Hinton]{maaten2008visualizing}
Maaten, L. v.~d. and Hinton, G.
\newblock Visualizing data using t-sne.
\newblock In \emph{JMLR}, 2008.

\bibitem[McCann et~al.(2017)McCann, Bradbury, Xiong, and
  Socher]{mccann-nips-2017}
McCann, B., Bradbury, J., Xiong, C., and Socher, R.
\newblock {Learned in translation: Contextualized word vectors}.
\newblock In \emph{NIPS}, 2017.

\bibitem[Melamud et~al.(2016)Melamud, Goldberger, and
  Dagan]{melamud-conll-2016}
Melamud, O., Goldberger, J., and Dagan, I.
\newblock {context2vec: Learning Generic Context Embedding with Bidirectional
  LSTM}.
\newblock In \emph{CoNLL}, 2016.

\bibitem[Perez et~al.(2018)Perez, Strub, De~Vries, Dumoulin, and
  Courville]{perez2018film}
Perez, E., Strub, F., De~Vries, H., Dumoulin, V., and Courville, A.
\newblock Film: Visual reasoning with a general conditioning layer.
\newblock In \emph{AAAI}, 2018.

\bibitem[Peters et~al.(2019)Peters, Ruder, and Smith]{peters2019tune}
Peters, M., Ruder, S., and Smith, N.~A.
\newblock To tune or not to tune? adapting pretrained representations to
  diverse tasks.
\newblock In \emph{arXiv:1903.05987}, 2019.

\bibitem[Peters et~al.(2018)Peters, Neumann, Iyyer, Gardner, Clark, Lee, and
  Zettlemoyer]{peters-arxiv-2018}
Peters, M.~E., Neumann, M., Iyyer, M., Gardner, M., Clark, C., Lee, K., and
  Zettlemoyer, L.
\newblock {Deep contextualized word representations}.
\newblock In \emph{NAACL}, 2018.

\bibitem[Pires et~al.(2019)Pires, Schlinger, and
  Garrette]{pires2019multilingual}
Pires, T., Schlinger, E., and Garrette, D.
\newblock How multilingual is multilingual bert?
\newblock In \emph{arXiv preprint arXiv:1906.01502}, 2019.

\bibitem[Puri \& Catanzaro(2019)Puri and Catanzaro]{puri2019zero}
Puri, R. and Catanzaro, B.
\newblock Zero-shot text classification with generative language models.
\newblock In \emph{arXiv:1912.10165}, 2019.

\bibitem[Radford et~al.(2018)Radford, Narasimhan, Salimans, and
  Sutskever]{radford2018improving}
Radford, A., Narasimhan, K., Salimans, T., and Sutskever, I.
\newblock {Improving language understanding by generative pre-training}.
\newblock In \emph{unpublished}, 2018.

\bibitem[Reimers \& Gurevych(2019)Reimers and Gurevych]{reimers2019sentence}
Reimers, N. and Gurevych, I.
\newblock Sentence-bert: Sentence embeddings using siamese bert-networks.
\newblock In \emph{EMNLP}, 2019.

\bibitem[Schwenk(2018)]{schwenk2018filtering}
Schwenk, H.
\newblock Filtering and mining parallel data in a joint multilingual space.
\newblock In \emph{arXiv:1805.09822}, 2018.

\bibitem[Schwenk et~al.(2019{\natexlab{a}})Schwenk, Kiela, and
  Douze]{schwenk2019analysis}
Schwenk, H., Kiela, D., and Douze, M.
\newblock Analysis of joint multilingual sentence representations and semantic
  k-nearest neighbor graphs.
\newblock In \emph{AAAI}, 2019{\natexlab{a}}.

\bibitem[Schwenk et~al.(2019{\natexlab{b}})Schwenk, Wenzek, Edunov, Grave, and
  Joulin]{schwenk2019ccmatrix}
Schwenk, H., Wenzek, G., Edunov, S., Grave, E., and Joulin, A.
\newblock Ccmatrix: Mining billions of high-quality parallel sentences on the
  web.
\newblock In \emph{arXiv:1911.04944}, 2019{\natexlab{b}}.

\bibitem[Shen et~al.(2019)Shen, Cheng, Sundararaman, Zhang, Yang, Tang,
  Celikyilmaz, and Carin]{shen2019learning}
Shen, D., Cheng, P., Sundararaman, D., Zhang, X., Yang, Q., Tang, M.,
  Celikyilmaz, A., and Carin, L.
\newblock Learning compressed sentence representations for on-device text
  processing.
\newblock In \emph{ACL}, 2019.

\bibitem[Stern et~al.(2019)Stern, Chan, Kiros, and
  Uszkoreit]{stern2019insertion}
Stern, M., Chan, W., Kiros, J., and Uszkoreit, J.
\newblock Insertion transformer: Flexible sequence generation via insertion
  operations.
\newblock In \emph{ICML}, 2019.

\bibitem[Stickland \& Murray(2019)Stickland and Murray]{stickland2019bert}
Stickland, A.~C. and Murray, I.
\newblock Bert and pals: Projected attention layers for efficient adaptation in
  multi-task learning.
\newblock In \emph{arXiv:1902.02671}, 2019.

\bibitem[Subramanian et~al.(2018)Subramanian, Trischler, Bengio, and
  Pal]{subramanian-iclr-2018}
Subramanian, S., Trischler, A., Bengio, Y., and Pal, C.
\newblock Learning general purpose distributed sentence representations via
  large scale multi-task learning.
\newblock In \emph{{ICLR}}, 2018.

\bibitem[van~den Oord et~al.(2016)van~den Oord, Kalchbrenner, Espeholt,
  Vinyals, Graves, et~al.]{van2016conditional}
van~den Oord, A., Kalchbrenner, N., Espeholt, L., Vinyals, O., Graves, A.,
  et~al.
\newblock {Conditional image generation with pixelcnn decoders}.
\newblock In \emph{NIPS}, 2016.

\bibitem[Vaswani et~al.(2017)Vaswani, Shazeer, Parmar, Uszkoreit, Jones, Gomez,
  Kaiser, and Polosukhin]{vaswani-nips-2017}
Vaswani, A., Shazeer, N., Parmar, N., Uszkoreit, J., Jones, L., Gomez, A.~N.,
  Kaiser, L., and Polosukhin, I.
\newblock {Attention Is All You Need}.
\newblock In \emph{{NIPS}}, 2017.

\bibitem[Wieting \& Kiela(2019)Wieting and Kiela]{wieting2019no}
Wieting, J. and Kiela, D.
\newblock No training required: Exploring random encoders for sentence
  classification.
\newblock In \emph{arXiv:1901.10444}, 2019.

\bibitem[Williams et~al.(2018)Williams, Nangia, and
  Bowman]{williams-naacl-2018}
Williams, A., Nangia, N., and Bowman, S.
\newblock {A Broad-Coverage Challenge Corpus for Sentence Understanding through
  Inference}.
\newblock In \emph{{NAACL}}, 2018.

\bibitem[Zweigenbaum et~al.(2018)Zweigenbaum, Sharoff, and
  Rapp]{zweigenbaum2018overview}
Zweigenbaum, P., Sharoff, S., and Rapp, R.
\newblock Overview of the third bucc shared task: Spotting parallel sentences
  in comparable corpora.
\newblock In \emph{Proceedings of 11th Workshop on Building and Using
  Comparable Corpora}, 2018.

\end{thebibliography}
\bibliographystyle{icml2020}

\end{document}